\documentclass{article}

\PassOptionsToPackage{numbers, compress}{natbib}
\usepackage[final]{tackling_climate_workshop_style}
\usepackage{graphicx}   
\usepackage{subcaption} 
\usepackage{float}
\usepackage{placeins}
\usepackage{array}
\usepackage{makecell}
\usepackage{tabularx}
\usepackage{caption}
\usepackage{adjustbox} 
\usepackage[utf8]{inputenc} 
\usepackage[T1]{fontenc}    
\usepackage{hyperref}       
\usepackage{url}            
\usepackage{booktabs}       
\usepackage{amsfonts}       
\usepackage{nicefrac}       
\usepackage{microtype}      
\usepackage{geometry} 
\usepackage{listings}   
\usepackage{xcolor}     
\usepackage{todonotes}
\usepackage{amsmath}
\newcommand{\model}{ALGae Observation and Segmentation}
\newcommand{\smodel}{ALGOS}

\title{Seg the HAB: Language-Guided Geospatial Algae Bloom Reasoning and Segmentation}
\author{%
Patterson Hsieh$^{1,}$\thanks{Equal Contribution}\qquad Jerry Yeh$^{2,}$\footnotemark[1]\qquad Mao-Chi He$^{2}$\footnotemark[1]  \qquad Wen-Han Hsieh$^{2,}$\qquad Elvis Hsieh$^{2,}$\\
\\ UC San Diego$^{1}$, UC Berkeley$^{2}$ \\
\texttt{\{h2hsieh\}@ucsd.edu} \quad \texttt{\{chiajui0721, matthew910818\}@gmail.com}
}

\begin{document}

\maketitle

\begin{abstract}
Climate change is intensifying the occurrence of harmful algal bloom (HAB), particularly cyanobacteria, which threaten aquatic ecosystems and human health through oxygen depletion, toxin release, and disruption of marine biodiversity. Traditional monitoring approaches, such as manual water sampling, remain labor-intensive and limited in spatial and temporal coverage. Recent advances in vision-language models (VLMs) for remote sensing have shown potential for scalable AI-driven solutions, yet challenges remain in reasoning over imagery and quantifying bloom severity. In this work, we introduce ALGae Observation and Segmentation (ALGOS), a segmentation-and-reasoning system for HAB monitoring that combines remote sensing image understanding with severity estimation. Our approach integrates GeoSAM-assisted human evaluation for high-quality segmentation mask curation and fine-tunes vision language model on severity prediction using the Cyanobacteria Aggregated Manual Labels (CAML) from NASA. Experiments demonstrate that ALGOS achieves robust performance on both segmentation and severity-level estimation, paving the way toward practical and automated cyanobacterial monitoring systems. The project website is available at: \url{https://patterson0128.github.io/hab.github.io/}.
\end{abstract}

\section{Introduction}
Harmful algal blooms (HAB) are an escalating global concern driven by climate change. Cyanobacteria-dominated HAB in particular present severe ecological, public health, and economic risks. Numerous studies have shown that HAB create multiple harms. From an ecological perspective, \citet{Anderson2021HABs} highlight that HAB cause mass fish mortality and ecosystem disruption. In addition to ecology, public health risks are also severe \citep{CDC2024HABIllness}. Economically, HAB impose billions of dollars in losses annually, specifically the 2017--2018 Florida Red Tide, causing an estimated \$2.7 million in losses \citep{NOAA2019RedTide,NOAA2025EconomicHABs}. These combined effects emphasize the idea that monitoring HAB dynamics is essential for mitigation and policy; however, existing methods, relying on sampling and manual microscopy, are costly, time-consuming, and geographically constrained.

Advances in computer vision and geospatial analysis provide new opportunities for scalable HAB monitoring. However, prior work in AI-based HAB monitoring has only tackled either bloom severity prediction or spatial segmentation, limiting comprehensive monitoring capabilities. Vision-based systems segment bloom patches in local camera imagery \citep{Barrientos2023}, yet they neither operate on wide-area remote-sensing data nor estimate severity, limiting the scalability. \citet{Dorne2024CyFi} estimates severity from Sentinel-2 and ancillary data but provides no explicit spatial delineation. Traditional remote-sensing methods using spectral indices can map blooms but depend on manual thresholds and site-specific tuning \citep{Yang2022Synoptic}. This fragmentation prevents systems from answering queries requiring both spatial and severity reasoning essential for targeted management. 

Motivated by these insights, we introduce {\smodel}, a unified vision-language framework that bridges reasoning segmentation with HAB severity assessment in satellite imagery. Following the prior work \citep{Quenum2025LISAT}, we leverage the CAML dataset for severity-level reasoning with a novel HAB segmentation dataset curated through GeoSAM-assisted annotation with human evaluation \citep{sultana2023geosam}. Our framework extends to HAB-specific remote sensing data, enabling simultaneous spatial localization of bloom extent and severity-level classification through natural language reasoning. Our experimental results demonstrate significant improvements over baseline segmentation models in spatial accuracy and baseline VLM in severity prediction. We believe {\smodel} explores a new path for automated HAB monitoring that combines the precision of pixel-level segmentation and the contextual reasoning capabilities necessary for ecological assessment and public health decision-making.

\section{Related Work}
\subsection{Cyanobacteria Detection and Segmentation}
Classical remote-sensing methods rely on spectral indices and thresholds, which can break down in optically complex inland waters and often require site-specific tuning \citep{Yang2022Synoptic}. Early computer-vision systems with hand-crafted color features were brittle to illumination and background changes \citep{Samantaray2018}. Modern deep models infer algal presence or chlorophyll-\emph{a} from multispectral imagery and often outperform traditional baselines, but generalization suffers when training data are region-limited \citep{Yang2022Synoptic}. 

Segmentation efforts show promise but are typically scoped to localization. \citet{Barrientos2023} segment CyanoHAB patches using synthetic imagery to mitigate data scarcity, yet do not infer severity. Conversely, multi-source fusion and ensemble models improve severity classification (e.g., Sentinel-2 + climate + terrain), but treat monitoring as point prediction without mapping spatial extent \citep{Nasios2025}. {\model} addresses these gaps by jointly producing a segmentation mask and a severity estimate with language-driven reasoning to support timely monitoring.

\subsection{Geospatial Foundation Models}
Geospatial foundation models extend vision--language models (VLMs) to remote sensing by pairing an overhead-image encoder with a language model so the system can describe scenes, answer questions, and follow instructions. Training the visual encoder on satellite or aerial imagery improves transfer. \citet{liu2024remoteclip}  aligns overhead images and text with CLIP-style contrastive learning. Instruction-tuned backbones (e.g., Vicuna) add conversational, task-following ability, but most current VLMs remain text-only, lacking spatial outputs such as maps or pixel-wise masks \citep{chiang2023vicuna}. To obtain spatially explicit results, recent work augments language models with lightweight segmentation branches so a query can return a pixel-level mask, and adapts generic segmenters to overhead data. For instance, a language-to-mask pathway has been explored in geospatial VLMs \citep{Quenum2025LISAT}, while GeoSAM fine-tunes Segment Anything for large, low-contrast satellite imagery \citep{sultana2023geosam,kirillov2023sam}. Together, these directions move geospatial AI from text answers toward actionable, per-pixel outputs that practitioners can use for monitoring and geospatial decision.

\section{Methods}
\begin{figure}
    \centering
    \includegraphics[width=1\linewidth]{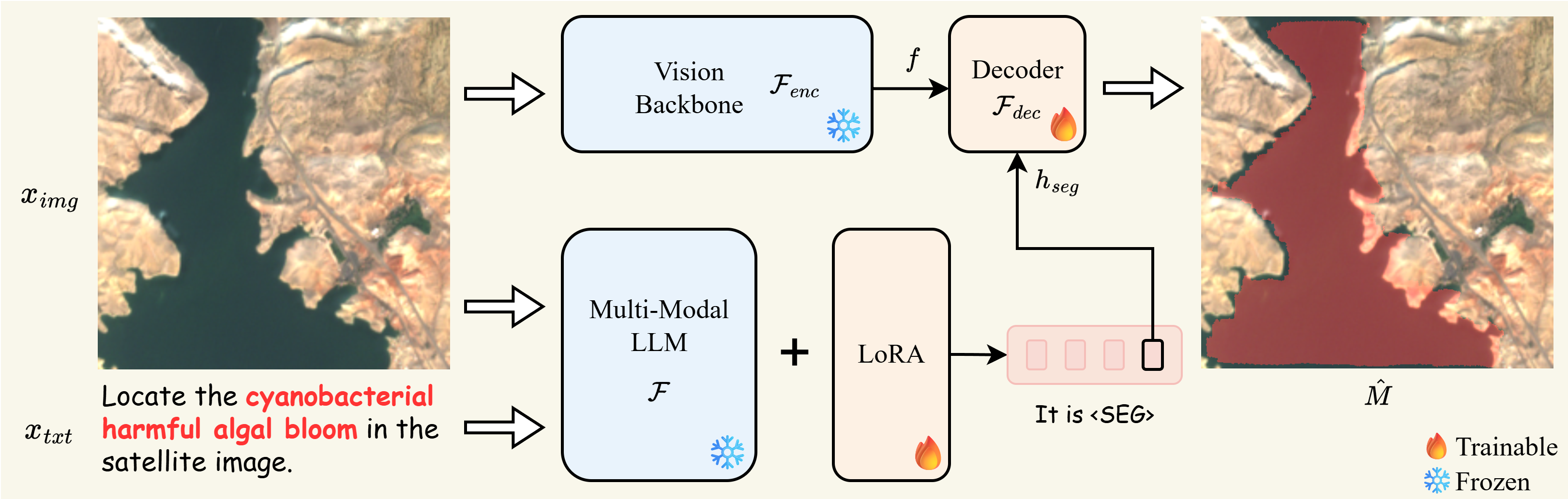}
    \caption{The pipeline of {\smodel}. Given the input image and text query, the multimodal LLM (\textsc{LISAt\textsubscript{pre}}) generates text output. The last-layer embedding for the <SEG> token is then decoded into the segmentation mask via the decoder. We adopt SAM \citep{kirillov2023sam} as our choice of vision backbone.}
    \label{fig:placeholder}
\end{figure}
\subsection{HAB Segmentation Dataset Curation}
Following the prior work \citep{Quenum2025LISAT}, we developed a comprehensive pipeline to generate high-quality pixel-level segmentation masks for HAB detection using the CAML dataset \citep{gupta2024caml}. Our approach addresses the unique challenges of HAB segmentation in Sentinel-2 imagery, where bloom boundaries are often diffuse with subtle spectral signatures.

We extend GeoSAM \citep{sultana2023geosam} with an interactive mask generation with human evaluation stage. Unlike the original GeoSAM, which primarily relies on automated point prompt generation, our framework introduces a semi-supervised curation loop. Users provide positive points on bloom areas, negative points on background regions, and a region-of-interest (ROI) box to guide the model’s attention. The system then generates candidate masks, which are interactively refined through lightweight morphological filtering and post-processing. This reduces ambiguity in diffuse bloom regions while retaining efficiency.

To ensure the reliability of the resulting dataset, each generated mask undergoes human evaluation. Annotators visually compare the candidate masks against the corresponding Sentinel-2 imagery, retaining only those masks that accurately delineate bloom regions. Masks that fail this validation are discarded or corrected through additional refinement. In this way, the semi-supervised feedback loop ensures that the segmentation masks are consistently aligned with ecological reality. Therefore, the resulting dataset contains high-quality segmentation masks, robust to the heterogeneous quality of Sentinel-2 imagery, from cloud-free high-resolution scenes to partially degraded scenes.

\subsection{HAB Reasoning Dataset Curation}
To enable severity-level assessment through natural language reasoning, we adopt the synthetic query generation pipeline following \citet{Quenum2025LISAT} to create HAB-reasoning queries.

\textbf{Severity-Based Query Generation.} We fine-grain the WHO recreational guidance thresholds \citep{who2003guidelines} into a five-level defined as follows: Level~1: $x < 2 \times 10^{4}$,  
Level~2: $2 \times 10^{4} \leq x < 1 \times 10^{5}$,  
Level~3: $1 \times 10^{5} \leq x < 1 \times 10^{6}$,  
Level~4: $1 \times 10^{6} \leq x < 1 \times 10^{7}$,  
Level~5: $x \geq 1 \times 10^{7}$ cells/mL. Based on the scale, we curate natural language query templates that require the model to infer algal bloom severity directly from satellite imagery. Each template prompts the model to classify severity on this ordinal five-point scale, ensuring consistent outputs while retaining ecological interpretability across heterogeneous observational conditions.

\textbf{Multi-modal Alignment.} Each reasoning query is paired with the corresponding satellite image and severity label defined above, creating a structured instruction–image–answer triplet (Appendix \ref{app:finetune}). This triplet design enables the model to jointly learn the relationship between visual appearance, ecological context, and ordinal severity categories, thereby supporting robust estimation of harmful algal blooms following \citep{Quenum2025LISAT, lai2024lisa}.

\subsection{Vision-Language Model Architecture for HAB Monitoring}
Our framework adopts the embedding-as-mask paradigm for HAB-specific applications, integrating domain-adapted visual encoders with language models fine-tuned on HAB-reasoning queries .

\textbf{Multimodal Integration.} Following LISAT's architecture \citep{Quenum2025LISAT}, we employ a Vicuna-7B language model \citep{chiang2023vicuna} as our base LLM, coupled with a Remote-CLIP ViT-L/14 encoder \citep{liu2024remoteclip} optimized for satellite imagery processing. The visual encoder processes Sentinel-2 multispectral bands, while a learnable linear projection aligns visual features with the language model's embedding space. We expand the vocabulary with a specialized \texttt{<SEG>} token that, when generated, triggers segmentation mask prediction through a SAM decoder head \citep{kirillov2023sam}. 

\textbf{Training Objectives.} 

Our model is optimized end-to-end with a joint objective that integrates both text generation and segmentation. The overall training loss $\mathcal{L}$ is formulated as a weighted combination:

\vspace{-.5cm}
\begin{equation}
    \mathcal{L} = \omega_{\text{txt}} \, \mathcal{L}_{\text{txt}} + \omega_{\text{mask}} \, \mathcal{L}_{\text{mask}}.
\end{equation}

Here, the text generation objective $\mathcal{L}_{\text{txt}}$ is defined as the standard autoregressive cross-entropy loss:
\vspace{-.1cm}
\begin{equation}
    \mathcal{L}_{\text{txt}} = \mathbf{CE}(\hat{\mathbf{y}}_{\text{txt}}, \mathbf{y}_{\text{txt}}).
\end{equation}

The segmentation objective $\mathcal{L}_{\text{mask}}$ combines a per-pixel binary cross-entropy (BCE) term and a DICE loss, balanced by $\omega_{\text{bce}}$ and $\omega_{\text{dice}}$:
\vspace{-.1cm}
\begin{equation}
    \mathcal{L}_{\text{mask}} = \omega_{\text{bce}} \, \mathbf{BCE}(\hat{\mathbf{M}}, \mathbf{M}) \;+\; \omega_{\text{dice}} \, \mathbf{DICE}(\hat{\mathbf{M}}, \mathbf{M}).
\end{equation}


\textbf{Implementation Details.}
All experiments were conducted on eight NVIDIA DGX A100 GPUs (80GB each). For severity prediction, we fine-tuned LLaVA-7B with LoRA for 50 epochs on the HAB dataset. For segmentation, ALGOS was trained jointly on HAB, FP-Ref-COCO \citep{wu2023seesaysegmentteaching}, and ReasonSeg \citep{lai2024lisa}. LoRA {\citep{hu2021loralowrankadaptationlarge}} was applied to the multimodal language model, while the SAM decoder was fully fine-tuned. The learning rate was set to $3 \times 10^{-4}$, with other configurations kept consistent with standard practice. For the composite loss, we set the weighting coefficients to $\omega_{\text{txt}} = 1.0$ and $\omega_{\text{mask}} = 1.0$. Within the segmentation objective, the per-pixel binary cross-entropy and DICE terms are weighted as $\omega_{\text{bce}} = 2.0$ and $\omega_{\text{dice}} = 0.5$, respectively. Empirically, this configuration yielded the best overall performance. Training required approximately 6 hours across eight DGX A100 GPUs.


\section{Results}
\subsection{Setup}
\textbf{Performance metrics.}
Following \citet{lai2024lisa,Quenum2025LISAT}, we evaluate segmentation using two IoU variants under a binary setting (algae vs.\ non-algae). The \textbf{cIoU} metric computes the per-image, class-balanced mean IoU; with only one foreground class, this reduces to the average IoU of algae masks across images (assigning a score of 1 when both prediction and ground truth are empty, and 0 otherwise). The \textbf{gIoU} metric instead aggregates intersections and unions across all test images before computing the ratio. In the binary case, this measures agreement with the \emph{total} algae extent over the full test set, making it more sensitive to prevalence and large contiguous blooms. For the severity prediction task, we use mean squared error (MSE) as the primary metric, reflecting the ordinal nature of severity levels.

\textbf{Baselines.} Table~\ref{tab:seg} compares {\smodel} with state-of-the-art reasoning segmentation models \citep{lai2024lisa,Quenum2025LISAT}, while Table~\ref{tab:llava_hab} benchmarks {\smodel} against LLaVA \citep{liu2023visualinstructiontuning}.

\subsection{Results and Observations}\label{sec:results}
{\smodel} achieves strong performance across both segmentation and severity prediction tasks. For segmentation, it reaches a cIoU of 0.65 and gIoU of 0.60, far surpassing LISAT (0.11 / 0.10) and LISA-7B (0.14 / 0.13), accurately capturing both per-image bloom regions and large contiguous extents across the dataset (Table~\ref{tab:seg}). For severity prediction, {\smodel} substantially reduces error, with mean squared error (MSE) dropping from 3.868 to 2.984, along with corresponding improvements in RMSE and MAE (Table~\ref{tab:llava_hab}). These results show that {\smodel} is capable of addressing both spatial segmentation and severity-level estimation, while outperforming baselines in each task.

\begin{table}
\begin{minipage}[t]{0.48\linewidth}
  \small
  \caption{Segmentation results comparing LISAT, LISA, and {\smodel}. cIoU: per-image class-balanced mean IoU; gIoU: dataset-level/global IoU.}
  \label{tab:seg}
  \setlength{\tabcolsep}{4pt}
  \begin{tabular}{lcc}
    \toprule
    \textbf{Model} & \textbf{cIoU} & \textbf{gIoU} \\
    \midrule
    LISAT     & $0.1083_{\pm 0.0124}$ & $0.1052_{\pm 0.0132}$ \\ 
    LISA 7B   & $0.1373_{\pm 0.0182}$ & $0.1274_{\pm 0.0160}$ \\
    \textbf{{\smodel}}  & $\textbf{0.6493}_{\pm 0.0301}$ & $\textbf{0.5969}_{\pm 0.0268}$ \\
    \bottomrule
  \end{tabular}
\end{minipage}
  \hfill
\begin{minipage}[t]{0.48\linewidth}
  \small
  \caption{Severity prediction results comparing the LLaVA baseline and {\smodel}.\\ \\}
  \label{tab:llava_hab}
  \setlength{\tabcolsep}{6pt}
  \begin{tabular}{lccc}
    \toprule
    \textbf{Model} & \textbf{MSE} & \textbf{RMSE} & \textbf{MAE} \\
    \midrule
    LLaVA-7B       & $3.868$ & $1.967$ & $1.587$ \\
    \textbf{{\smodel}} & $\textbf{2.984}$ & $\textbf{1.727}$ & $\textbf{1.365}$ \\
    \bottomrule
  \end{tabular}
\end{minipage}
\end{table}

\section{Conclusion}
We introduced {\model}, a framework that leverages multimodal language models to reason over heterogeneous data and dynamically segment harmful algal bloom (HAB) regions. Additionally, we proposed a semi-supervised segmentation pipeline that improves delineation in images where automated methods struggle with unclear bloom boundaries. {\smodel} synthesizes spatial and contextual information through a structured reasoning process, ensuring that the segmented regions align with bloom severity levels. Unlike prior work, which has only addressed either severity estimation or localized segmentation in isolation, our approach integrates geospatial foundation models to jointly perform both tasks on wide-area remote sensing imagery. By advancing the ability of geospatial models to reason and adaptively segment polluted areas, {\smodel} provides a robust tool for ecological monitoring and policy support, enabling scalable HAB monitoring.

\textbf{Limitations.} Our framework has been evaluated on a limited geographic and seasonal scope based on the CAML dataset, and its generalization to diverse aquatic environments requires larger-scale, cross-region benchmarks. In addition, the reliance on curated datasets highlights the need for continuous data integration that can adapt to evolving ecological conditions. In future work, we will address both limitations by extending our evaluations and data pipelines to support scalable deployment.



\bibliography{custom}

\appendix
\clearpage
\FloatBarrier
\setlength{\tabcolsep}{5pt}
\renewcommand{\arraystretch}{1.12}
\captionsetup[table]{skip=6pt}
\captionsetup[figure]{skip=6pt}
\newcolumntype{Q}{>{\raggedright\arraybackslash}m{0.24\textwidth}}
\newcolumntype{Y}{>{\centering\arraybackslash}m{0.18\textwidth}}
\newcommand{\imgh}{3.0cm}
\newcommand{\queryrow}[5]{%
  #1 &
  \includegraphics[width=\linewidth]{#2} &
  \includegraphics[width=\linewidth]{#3} &
  \includegraphics[width=\linewidth]{#4} &
  \includegraphics[width=\linewidth]{#5} \\
}
\setcounter{section}{0} 

\section{Qualitative Comparison Across Models}
\label{app:qualitative}

\begin{center}
\begin{minipage}{\textwidth}
\centering

\begin{adjustbox}{width=\textwidth}  
\begin{tabular}{QYYYY}
\toprule
\textbf{Queries} & \textbf{LISA--7B} & \textbf{LISAT} & \textbf{ALGOS (Ours)} & \textbf{Ground Truth} \\
\midrule

\queryrow{Locate the cyanobacterial harmful algal bloom in the satellite image.}
{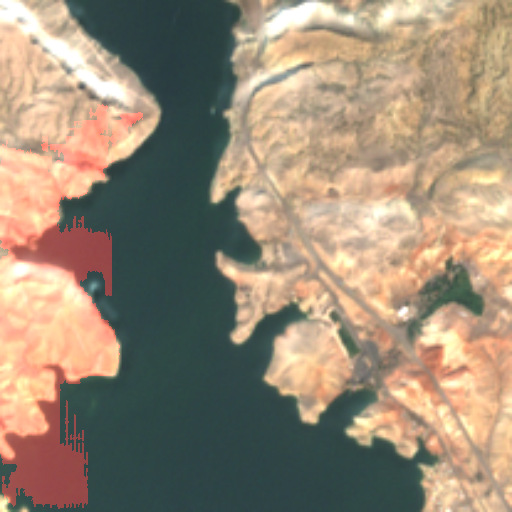}
{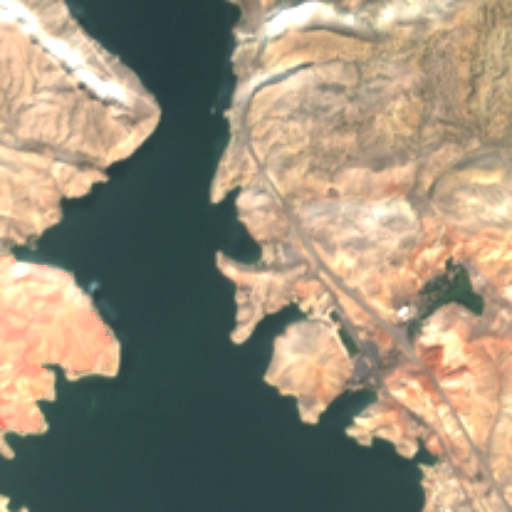}
{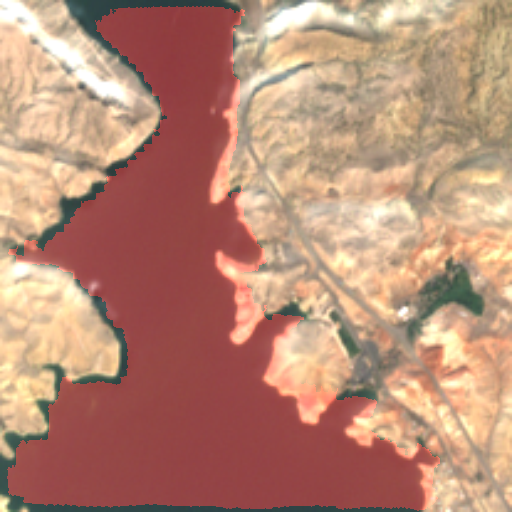}
{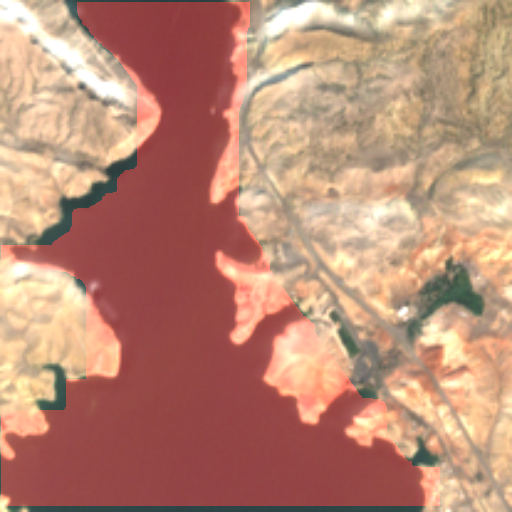}

\queryrow{Locate all visible harmful algal blooms.}
{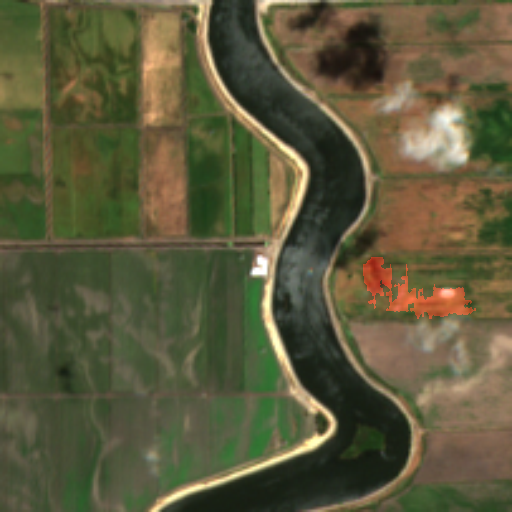}
{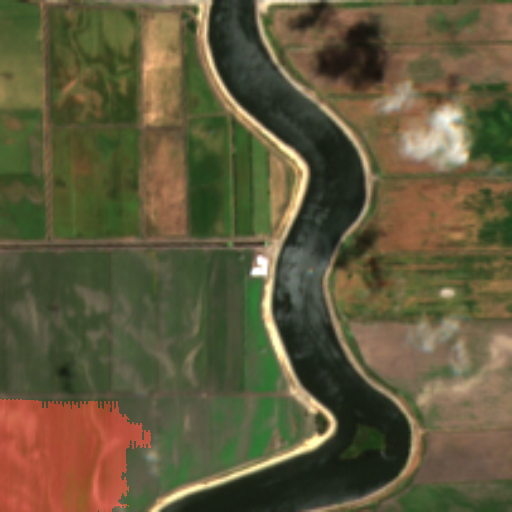}
{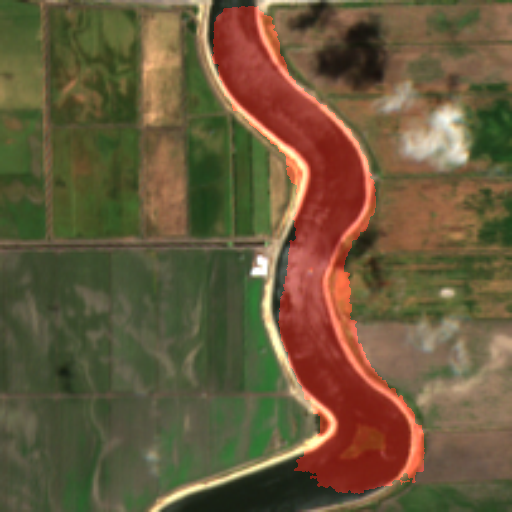}
{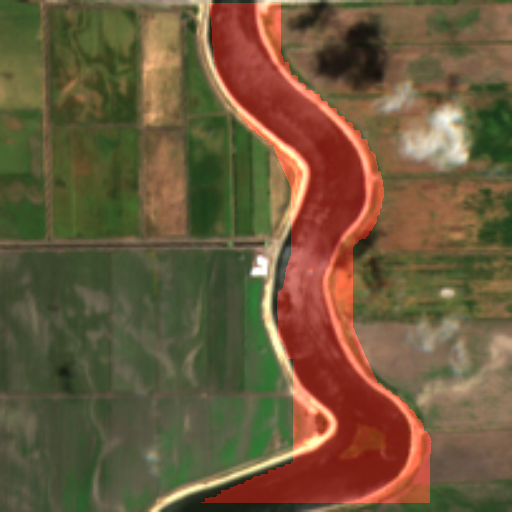}

\queryrow{Find the cyanobacterial harmful algal bloom in the satellite image.}
{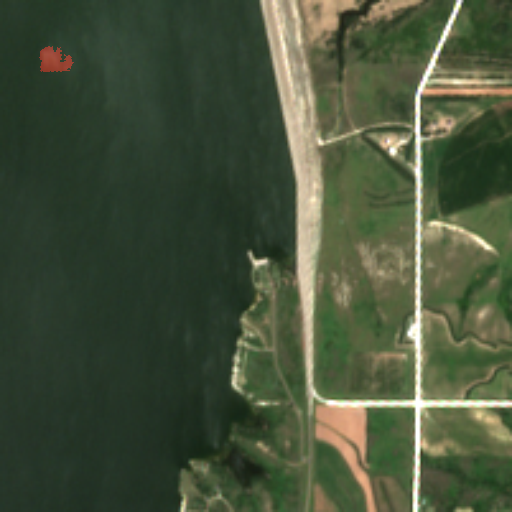}
{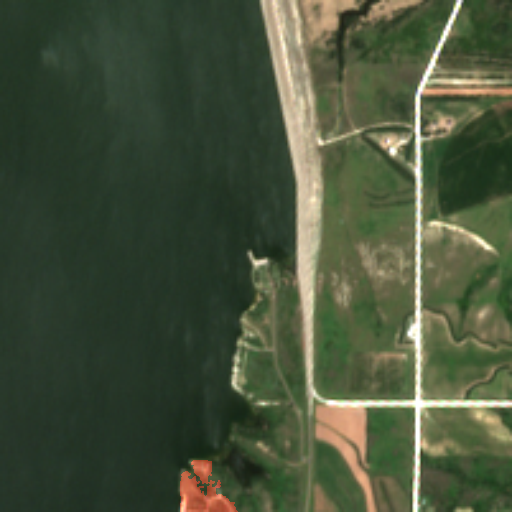}
{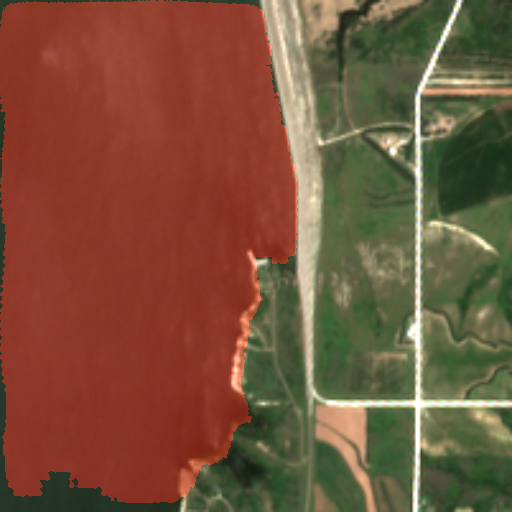}
{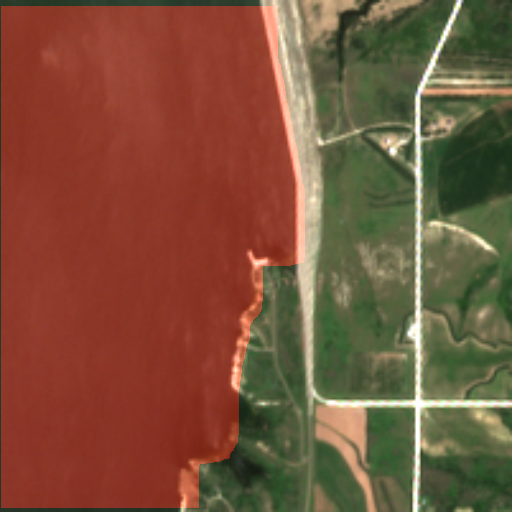}

\queryrow{Segment all visible harmful algal blooms.}
{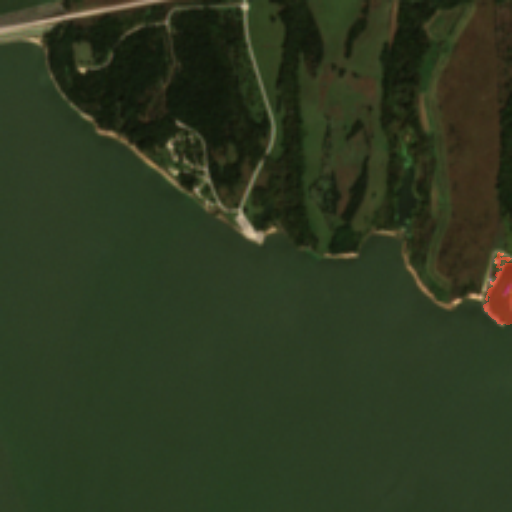}
{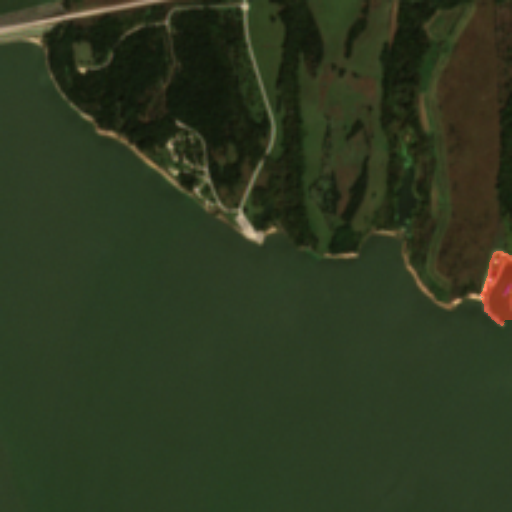}
{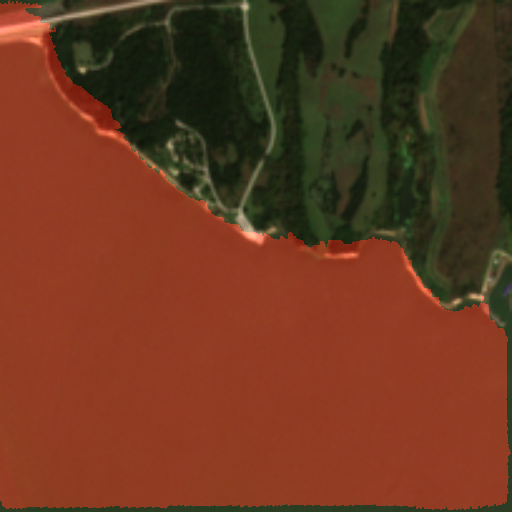}
{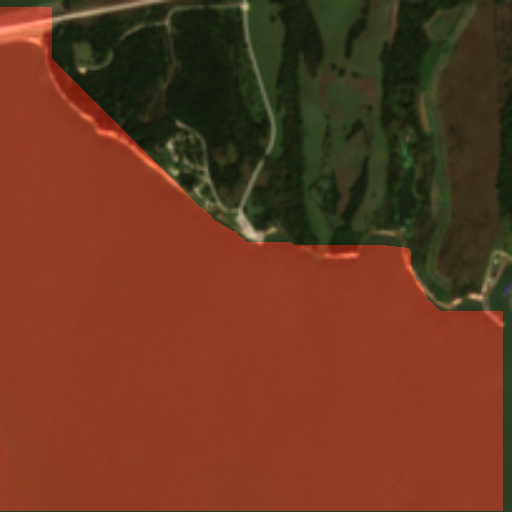}

\queryrow{Segment the waterbody affected by cyanobacteria.}
{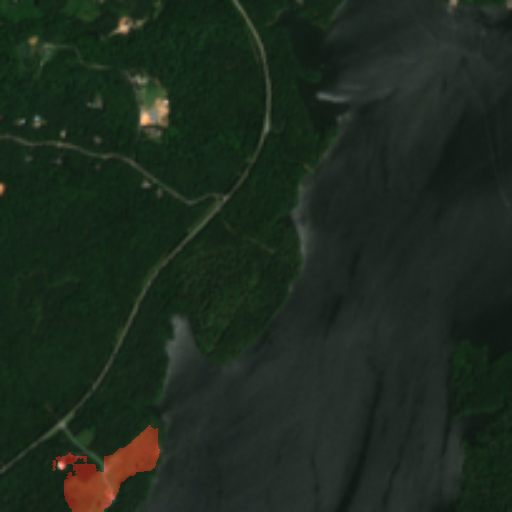}
{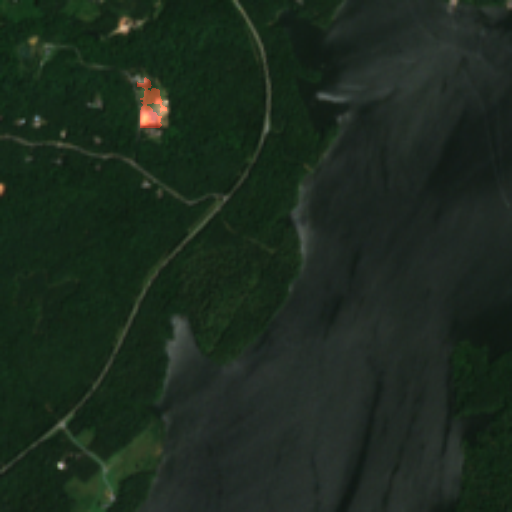}
{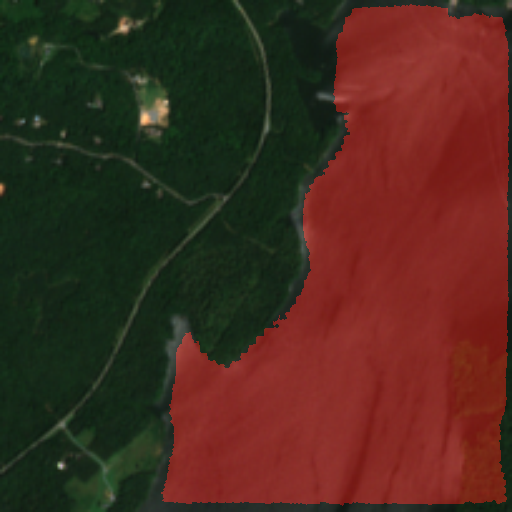}
{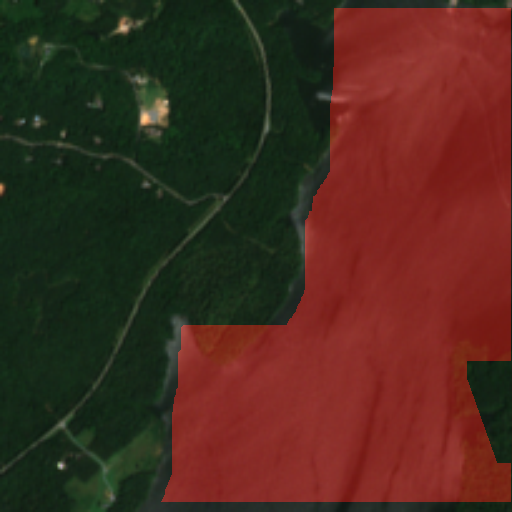}

\queryrow{Segment the cyanobacterial harmful algal bloom in the satellite image.}
{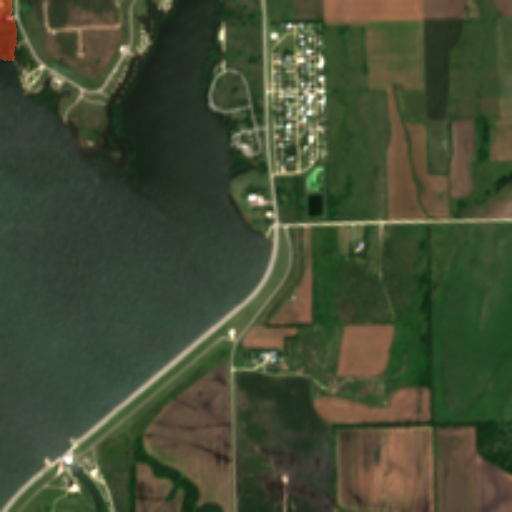}
{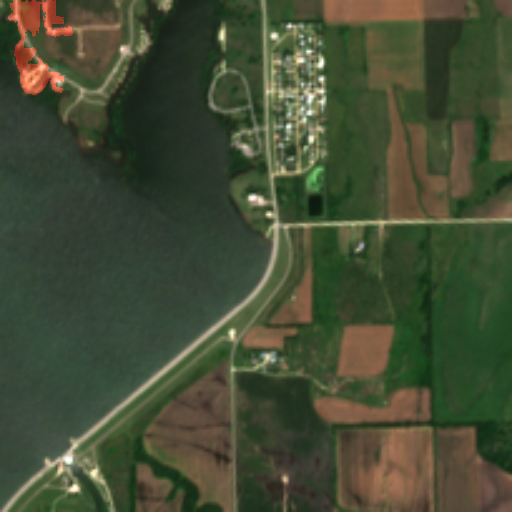}
{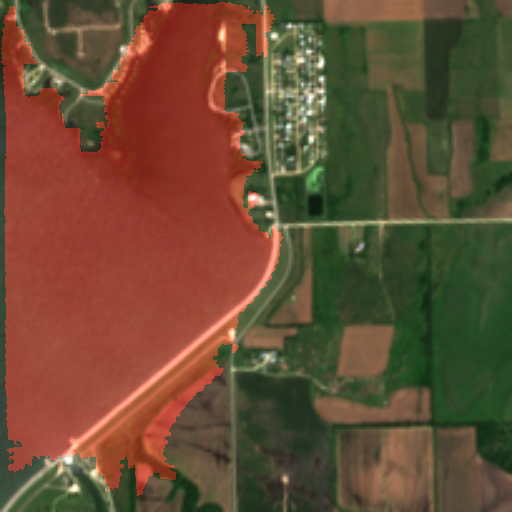}
{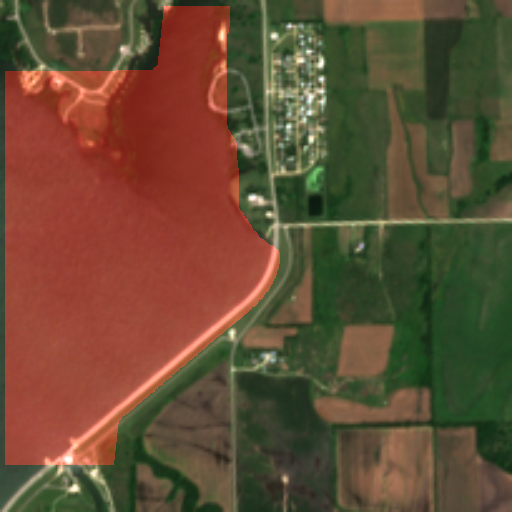}

\queryrow{Segment the algal bloom affected areas.}
{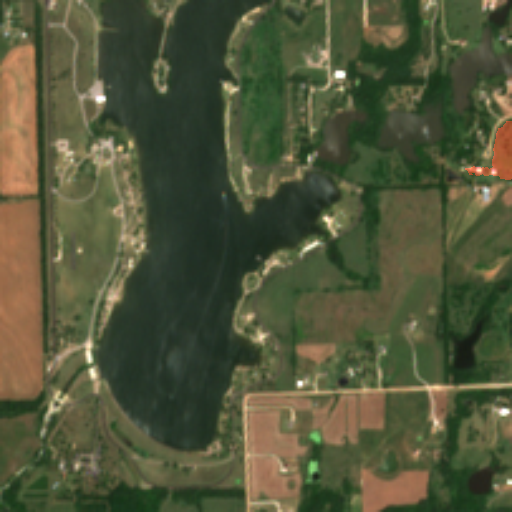}
{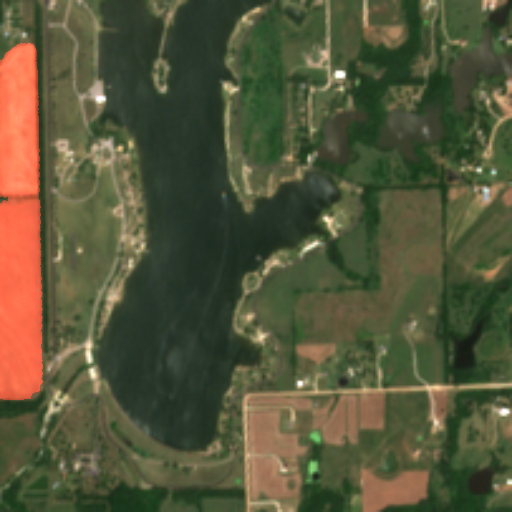}
{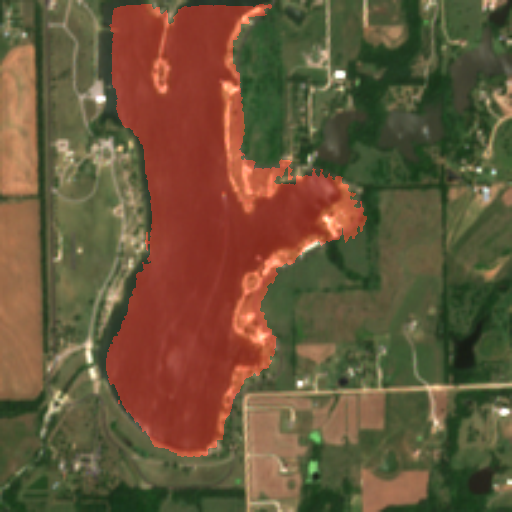}
{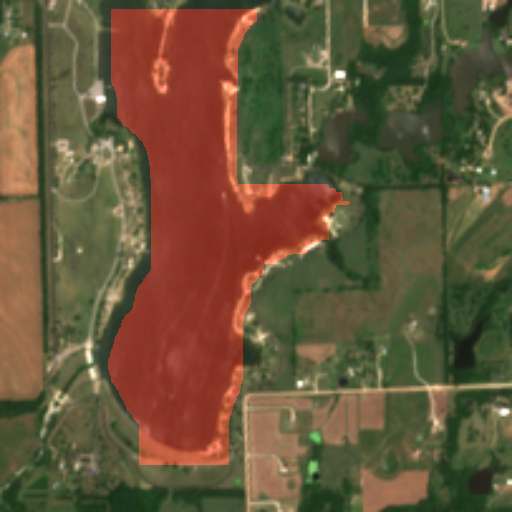}

\queryrow{Locate the cyanobacterial harmful algal bloom in the satellite image.}
{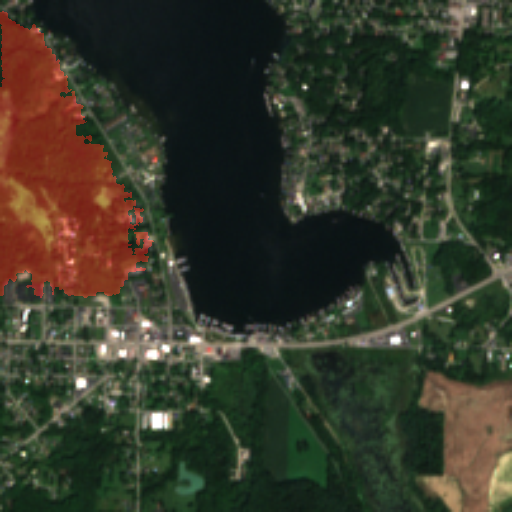}
{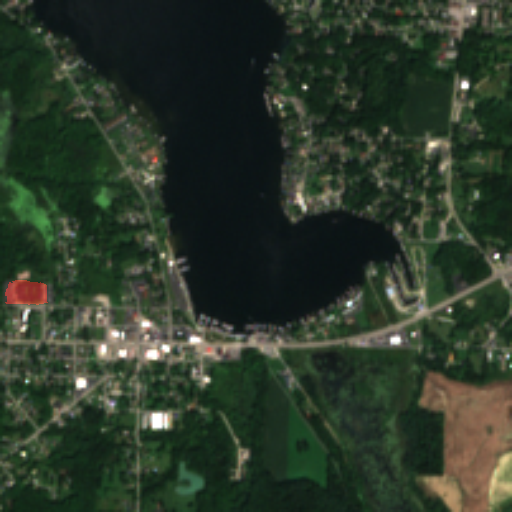}
{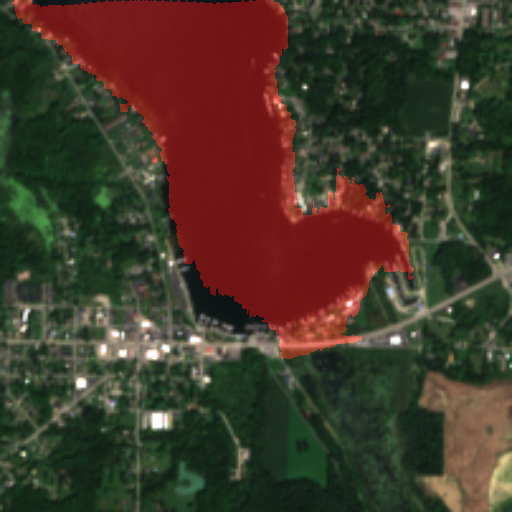}
{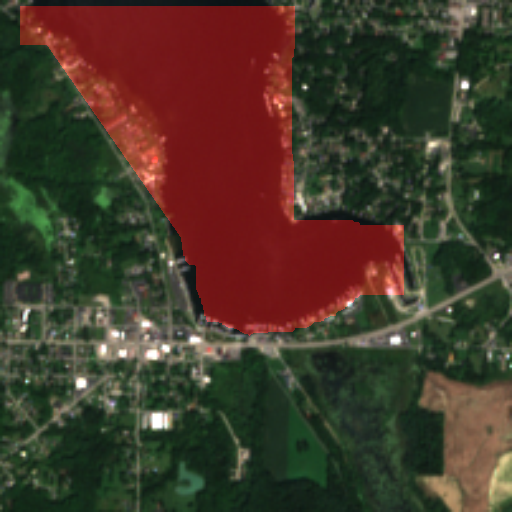}

\bottomrule
\end{tabular}
\end{adjustbox}

\label{tab:qualitative_grid}
\end{minipage}
\end{center}

\captionof{figure}{Qualitative comparison of predictions across models.}

\clearpage
\appendix
\setcounter{section}{1} 
\section{Image Segmentation}
\label{app:segmentation}

\begin{figure}[H]
    \centering

    \includegraphics[width=0.32\textwidth]{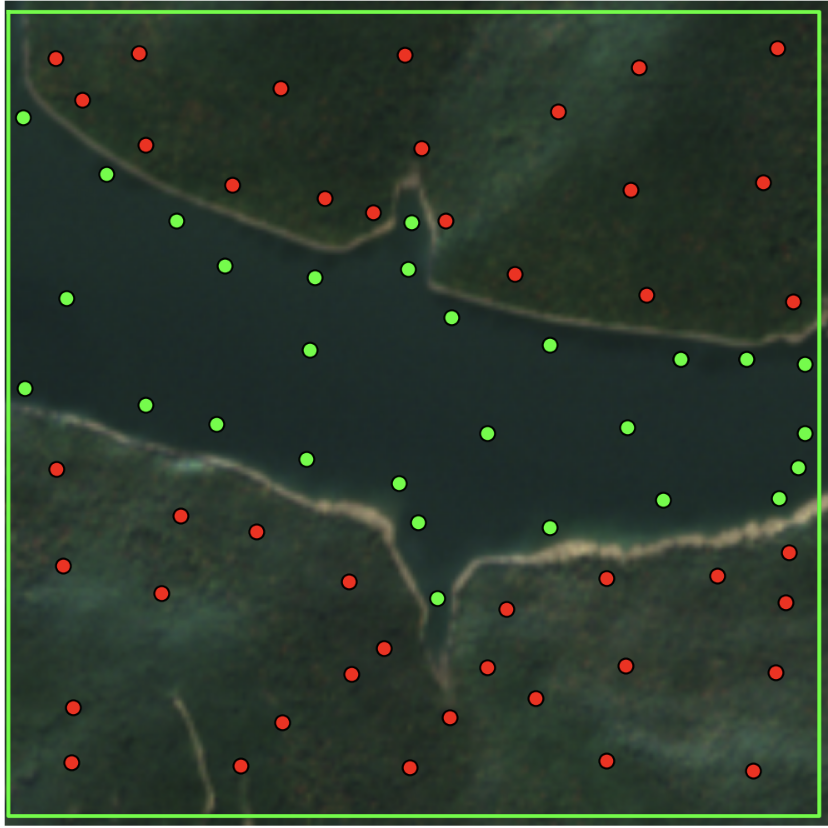}
    \includegraphics[width=0.32\textwidth]{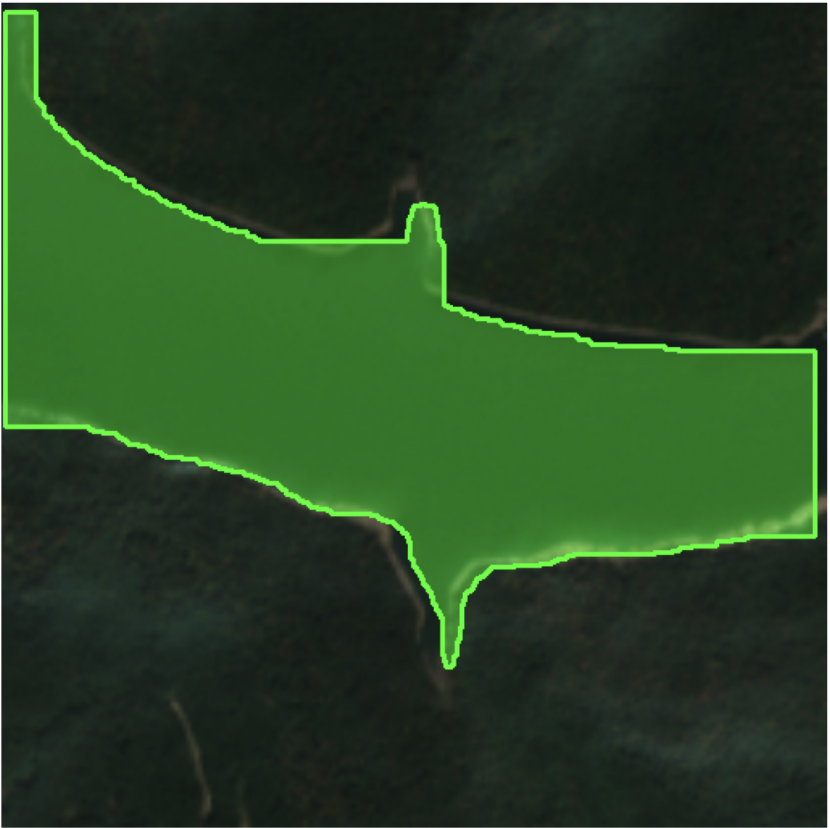}
    \includegraphics[width=0.32\textwidth]{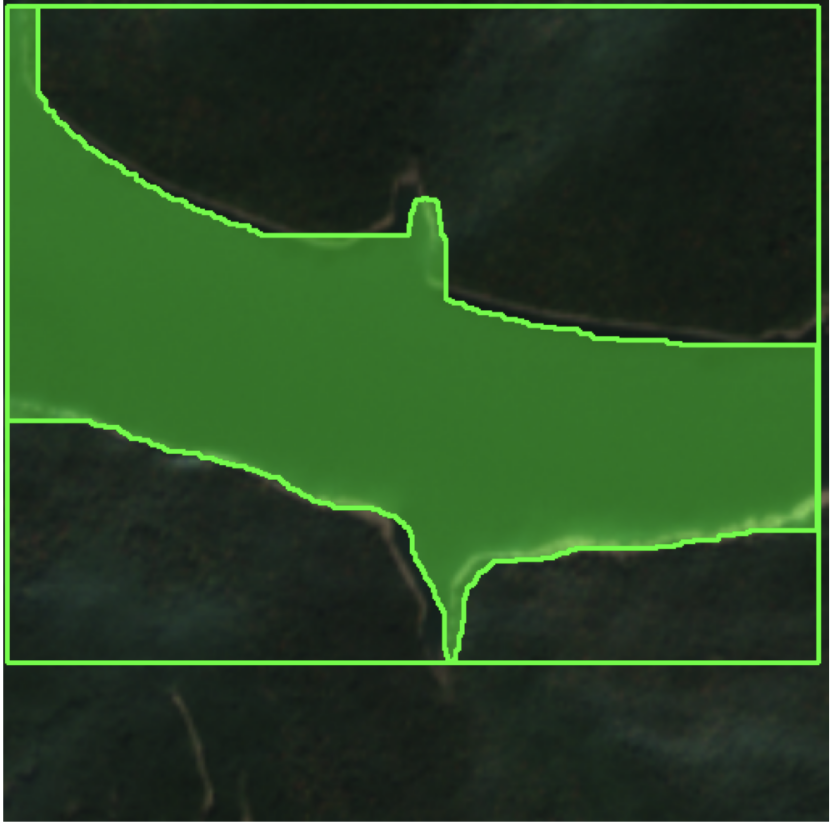}

    \includegraphics[width=0.32\textwidth]{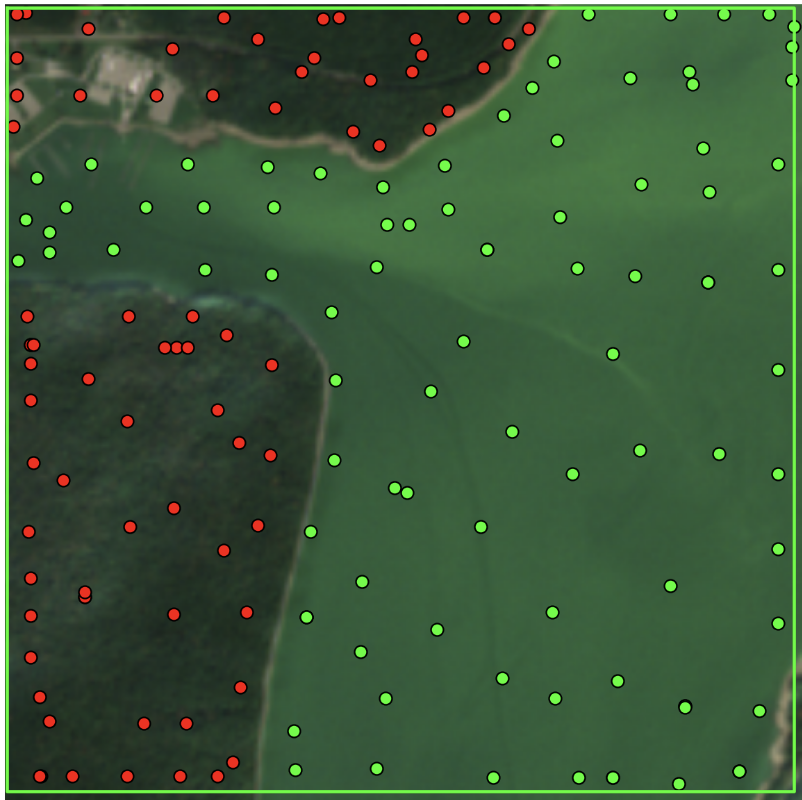}
    \includegraphics[width=0.32\textwidth]{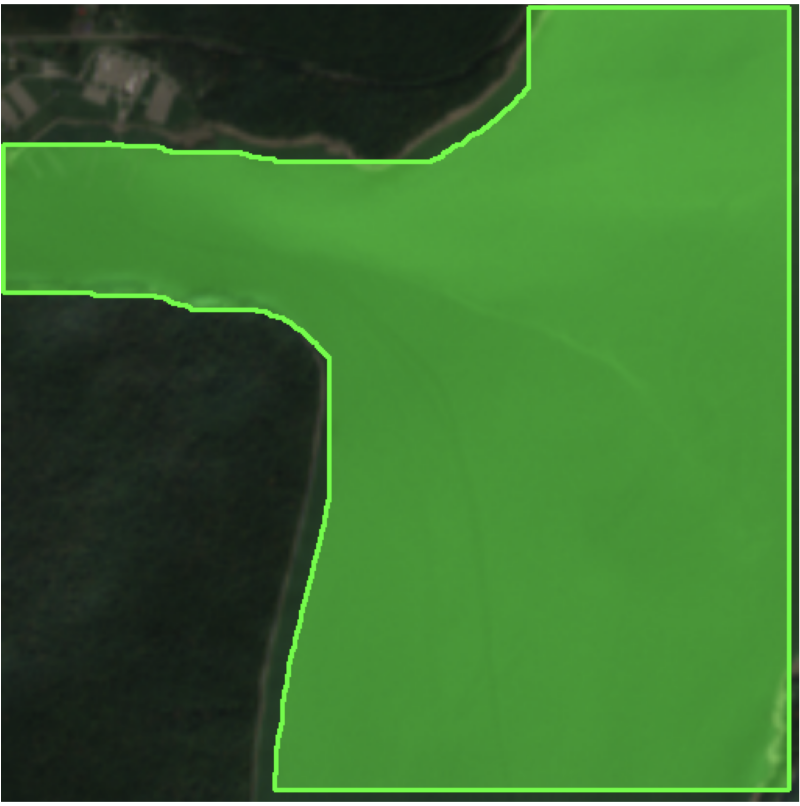}
    \includegraphics[width=0.32\textwidth]{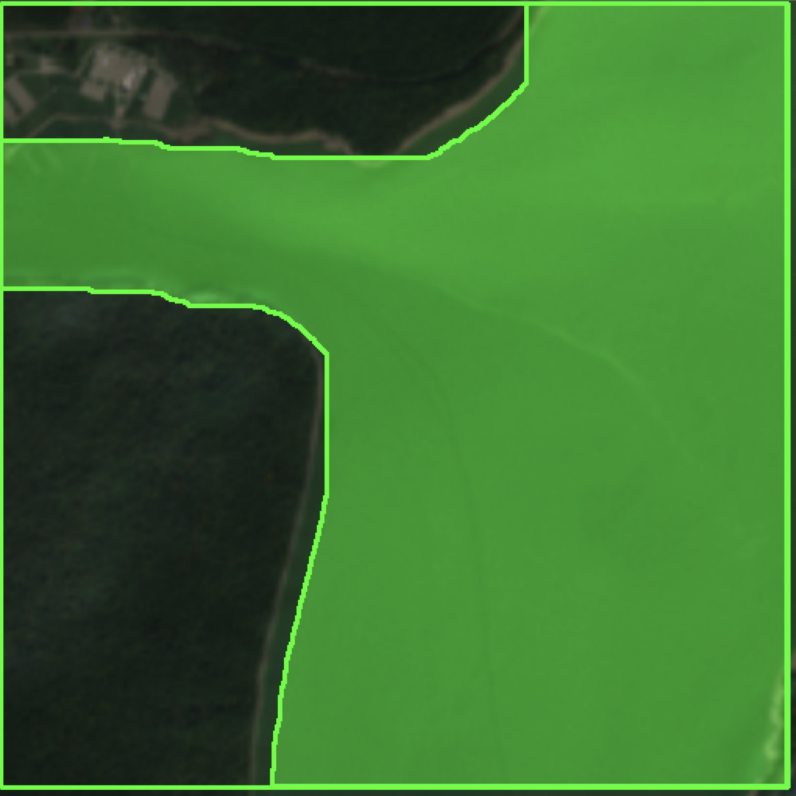}

    \begin{subfigure}[b]{0.32\textwidth}
        \centering
        \includegraphics[width=\textwidth]{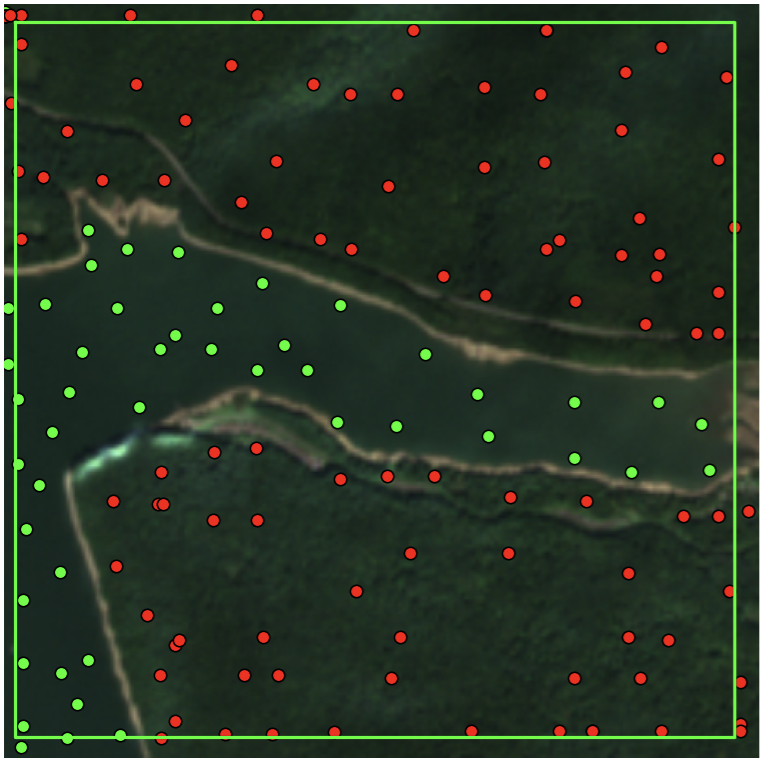}
        \caption{User-provided prompts.}
    \end{subfigure}
    \begin{subfigure}[b]{0.32\textwidth}
        \centering
        \includegraphics[width=\textwidth]{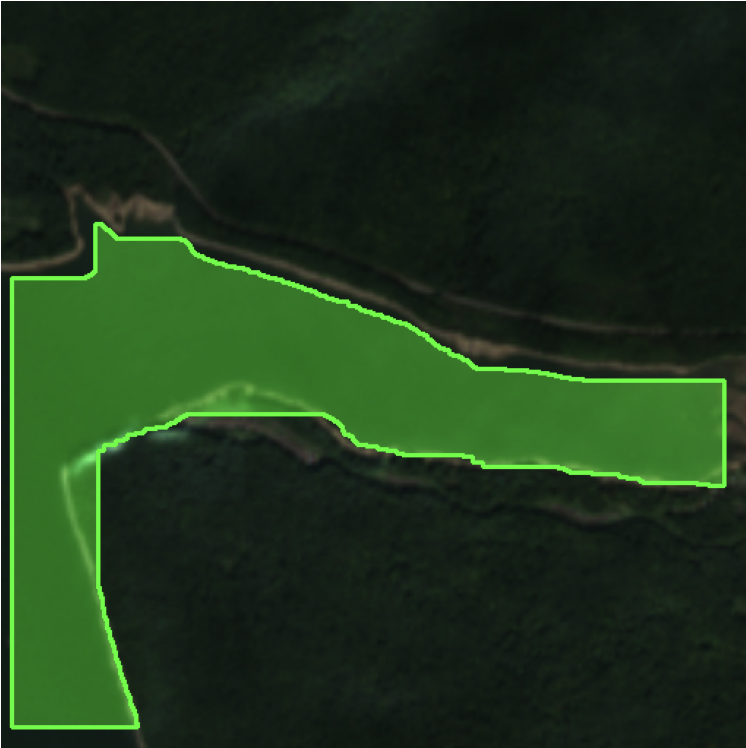}
        \caption{Generated mask overlay.}
    \end{subfigure}
    \begin{subfigure}[b]{0.32\textwidth}
        \centering
        \includegraphics[width=\textwidth]{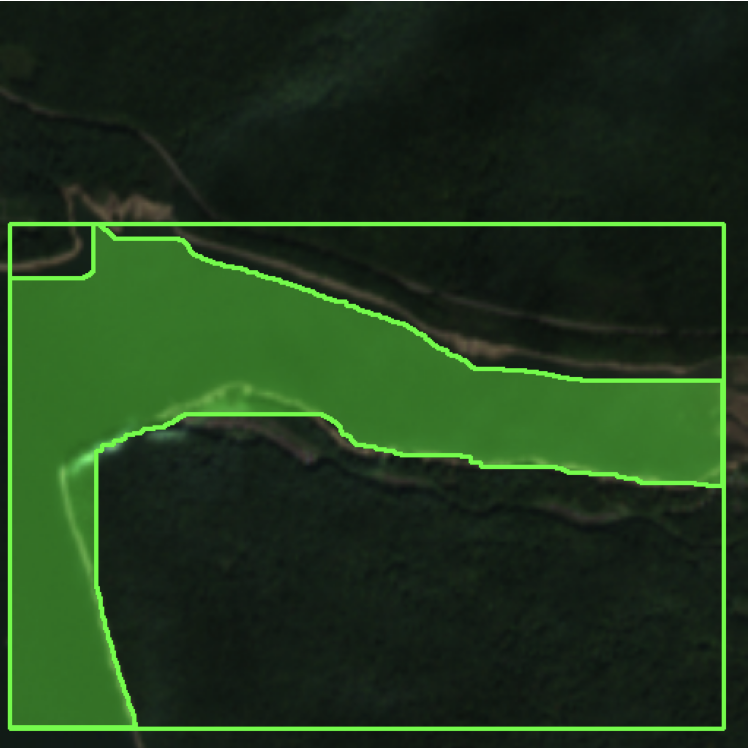}
        \caption{Bounding boxes extracted.}
    \end{subfigure}

    \caption{Interactive segmentation workflow used during data curation.}
    \label{fig:interactive_segmentation}
\end{figure}

In this section, we illustrate the interactive segmentation workflow used for data curation 
(Figure~\ref{fig:interactive_segmentation}). Users first provide positive (green) and negative (red) 
prompts to guide the model. The model then generates a segmentation mask on Sentinel-2 imagery, 
from which bounding boxes are extracted to obtain object-level representations for downstream analysis.

\clearpage
\appendix
\setcounter{section}{2} 
\section{Algae Severity Assessment}
\label{app:finetune}
\definecolor{codegray}{gray}{0.95}
\lstset{
  backgroundcolor=\color{codegray},
  basicstyle=\ttfamily\small,
  breakatwhitespace=false,
  breaklines=true,
  captionpos=b,
  frame=single,
  rulecolor=\color{black!30},
  showspaces=false,
  showstringspaces=false,
  showtabs=false,
  tabsize=2
}

This sample demonstrates a structured \textit{instruction–image–answer} triplet used for model fine-tuning.  
Severity levels follow the WHO recreational thresholds, refined into five ordinal categories:  
1 = Very low, 2 = Low, 3 = Moderate, 4 = High, 5 = Very high.  

\subsection*{Query Prompt}
\begin{lstlisting}[language={}]
<image>
Analyze the provided satellite image of algae-specific conditions. Determine the severity level, where: 
1 = Very low, 2 = Low, 3 = Moderate, 4 = High, 5 = Very high.
Output only a single digit from 1-5 with no other text.
Example output:3
\end{lstlisting}

\subsection*{Input Image}
\begin{figure}[H]
    \centering
    \includegraphics[width=0.7\textwidth]{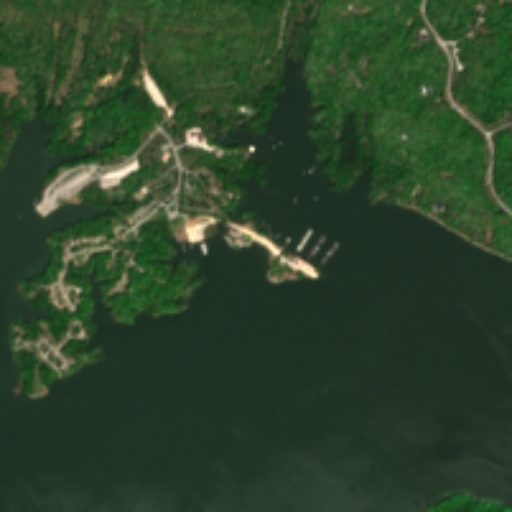}
    \caption{Satellite image provided as input.}
    \label{fig:sample1}
\end{figure}

\subsection*{Ground Truth Label}
This example corresponds to \textbf{Level 1 (very low)}, with an expected output of: \texttt{1.0}

\end{document}